\documentclass[letterpaper, 10 pt, conference]{ieeeconf}  
\IEEEoverridecommandlockouts 
\usepackage{amsmath,amsfonts}
\usepackage{algorithmic}
\usepackage{algorithm}
\usepackage{array}
\usepackage[caption=false,font=normalsize,labelfont=sf,textfont=sf]{subfig}
\usepackage{textcomp}
\usepackage{stfloats}
\usepackage{url}
\usepackage{verbatim}
\usepackage{graphicx}
\usepackage{newtxmath}
\usepackage{cite} 
\title{\LARGE \bf
CrackSegDiff: Diffusion Probability Model-based Multi-modal Crack Segmentation}
\author{Xiaoyan Jiang$^{1*}$, Licheng Jiang$^{1*}$, Anjie Wang$^{2}$, Kaiying Zhu$^{3}$, Yongbin Gao$^{1\dagger}$
\thanks{* Equal contribution.}
\thanks{$\dagger$ Corresponding author: {\tt\small gaoyongbin@sues.edu.cn}}
\thanks{$^{1}$ School of Electronic and Electrical Engineering, Shanghai University of Engineering Science.}
\thanks{$^{3}$ School of Electronic and Computer Engineering, Peking University.}
\thanks{$^{4}$ SenseTime.}
}%

\begin{document}
\maketitle
\thispagestyle{empty}
\pagestyle{empty}
\begin{abstract}
Integrating grayscale and depth data in road inspection robots could enhance the accuracy, reliability, and comprehensiveness of road condition assessments, leading to improved maintenance strategies and safer infrastructure. However, these data sources are often compromised by significant background noise from the pavement. Recent advancements in Diffusion Probabilistic Models (DPM) have demonstrated remarkable success in image segmentation tasks, showcasing potent denoising capabilities, as evidenced in studies like SegDiff \cite{amit2021segdiff}. Despite these advancements, current DPM-based segmentors do not fully capitalize on the potential of original image data. In this paper, we propose a novel DPM-based approach for crack segmentation, named CrackSegDiff, which uniquely fuses grayscale and range/depth images. This method enhances the reverse diffusion process by intensifying the interaction between local feature extraction via DPM and global feature extraction. Unlike traditional methods that utilize Transformers for global features, our approach employs Vm-unet \cite{ruan2024vm} to efficiently capture long-range information of the original data. The integration of features is further refined through two innovative modules: the Channel Fusion Module (CFM) and the Shallow Feature Compensation Module (SFCM). Our experimental evaluation on the three-class crack image segmentation tasks within the FIND dataset demonstrates that CrackSegDiff outperforms state-of-the-art methods, particularly excelling in the detection of shallow cracks.
Code is available at https://github.com/sky-visionX/CrackSegDiff.
\end{abstract}

\section{Introduction}
Nowadays, road inspection robots equipping multiple sensors are adopted worldwidely for road structure health monitoring and condition assessment.
Among the defects that affect the road health condition, cracks are the most common but challenging type to be detected.
Crack segmentation involves identifying cracks on a pavement image at the pixel level, providing accurate shapes and locations of cracks as feedback to the robots.
Most machine vision-based crack segmentation algorithms are based on neural network models, for instance, convolutional neural networks (CNN)-based CrackNet-V \cite{mei2020multi}, generative adversarial networks (GAN)-based CrackGAN \cite{zhang2020crackgan}, and SCDeepLab \cite{zhou2023hybrid} combining CNN and Transformer \cite{vaswani2017attention}.
\begin{figure}[htb]
\centering{\includegraphics[width=0.48\textwidth]{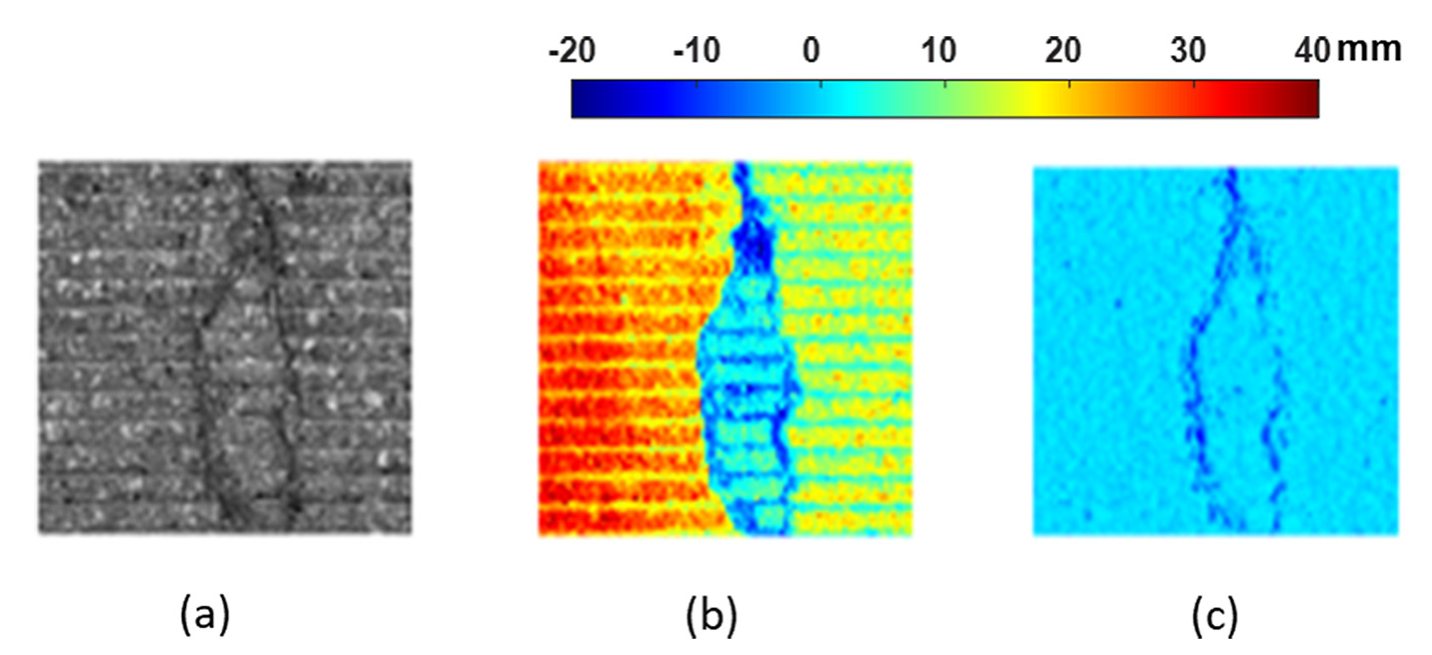}%
\label{fig_first_case}}
\caption{Example images captured from roadway surface \cite{zhou2023deep}: (a) raw intensity image; (b) raw range image; and (c) filtered range image.
}
\label{fig_sim}
\end{figure}
The presence of background clutters and the varied appearances of cracks in pavement images pose significant challenges for accurate crack segmentation in practical applications. 
This variability often leads to misidentification, where cracks are mistakenly recognized as scratches, water streaks, or tar lines. 
As Fig. 1 shows, depth or range data could provide more information to enhance reliability and yield more consistent results across different environments.

Recently, Diffusion Probabilistic Models have gained much attention and great success in generative tasks, without absolute ground truth. 
Sequentially, multiple DPM-based segmentors, such as SegDiff \cite{amit2021segdiff} and MedSegDiff \cite{wu2024medsegdiff}, are introduced to image segmentation using the powerful denoising ability of DPM conditioned on input image. 
Intuitively, DPM-based segmentor is suitable for the task at hand. 
But, the reverse diffusion process of the diffusion model in SegDiff has not effectively utilized the information from the original image to guide the denoising process.
Noteworthy, due to noise and background variations in pavement images, DPM's U-Net backbone losts crack structural information as the network depth increases during training. 
This makes DPM-based segmentors insensitive to contextual information, which is crucial for crack segmentation. 
Hence, Transformer is integrated as the global features to supplement the local features extracted by DPM \cite{chowdary2023diffusion, zhu2024diffswintr, liu2024transdiff}.

However, directly implementing this approach led to poor performance \cite{wu2024medsegdiff}. 
One issue is the incompatibility between the abstract conditional features of the Transformer and the features of the diffusion backbone. 
The Transformer learns deep semantic features from the original image, while the diffusion backbone extracts features from the original image masked with Gaussian noise, making feature fusion more challenging. 
Additionally, the Transformer's computational complexity, dynamism, and globality make it more sensitive than CNN.
Recently, state-space models (SSM) represented by Mamba \cite{gu2023mamba} not only establish long-range dependencies but also maintain linear computational complexity. 
Vm-unet \cite{ruan2024vm} is a U-shaped architecture model using SSM for image segmentation, capable of capturing extensive global information and compatible with the U-Net backbone of the diffusion model.

In this paper, we introduce an optimized DPM-based framework for crack segmentation, which integrates grayscale imagery and range data. Utilizing SegDiff \cite{amit2021segdiff} as the backbone and a feature enhancement module composed of Vm-unet \cite{ruan2024vm}, this model effectively captures long-range dependencies and integrates global features into the U-Net architecture, addressing the lack of global context in SegDiff.

To overcome the challenge of feature compatibility and enhance performance, we have developed the Channel Fusion Module (CFM). This module synergizes multi-scale global and local features from both grayscale and depth images, harnessing their complementary strengths. Grayscale images provide texture details, while depth images offer structural insights, and CFM facilitates their integration through spatial and channel fusion, maximizing the utility of both data types.

Furthermore, detecting shallow cracks remains a challenge due to the loss of low-level features in deeper network layers, exacerbated by noise and background variations in pavement images. To address this, we propose the Shallow Feature Compensation Module (SFCM), which is designed to preserve these critical low-level features, thereby enhancing the segmentation of shallow cracks.

Contributions are summarized as follows:
1) We pioneer the application of a diffusion model for crack segmentation, employing the Feature Enhancement Model (FEM) with Vm-unet for robust global feature integration and enhanced through mixed-loss supervision.
2) We introduce the Shallow Feature Compensation Module, which preserves structural features of cracks by enriching high-level features with multi-scale low-level details, thereby enhancing segmentation accuracy and reducing noise interference.
3) We develop the Channel Fusion Module, designed to seamlessly integrate and optimize the synergy between multi-scale global and local features extracted from both grayscale and depth images.
4) Our CrackSegDiff framework sets a new benchmark for state-of-the-art performance on the FIND dataset across various modalities, including grayscale images, depth images, and their fusion, demonstrating superior crack image segmentation capabilities.
 
\section{Related Work}
\subsection{Deep Learning-based Crack Segmentation}
Traditionally, crack segmentation relies on CNN to predict the classification label for each pixel. While CNN used in studies, such as \cite{di2023u}, \cite{liu2019deepcrack}, and \cite{fei2019pixel}, provide reliable baseline performance, they inherently struggle to capture long-range dependencies. 
This limitation often leads to discontinuous crack detection and false segmentations in complex environments.
To address these issues, recent works like \cite{xiang2023crack}, \cite{wang2024swincrack}, and \cite{quan2023crackvit} offer more robust solutions by leveraging both local feature and global context modeling.
They combine CNN with Transformers, introducing specialized modules to better balance the fusion of local and global features, mitigating the drawbacks of purely CNN-based models.

GAN have also been explored in crack segmentation, offering a way to generate finer details, such as crack boundaries. Studies like \cite{gao2019generative} and \cite{9770464} demonstrate the potential of GAN to handle noise and complex backgrounds effectively. However, GAN are often hindered by the implicit nature of their learning process, leading to challenges like unstable training, mode collapse, and artifacts. Research such as \cite{tian2021new} and \cite{pan2023automatic} addresses these limitations by employing techniques like penalty terms and joint loss functions to improve performance.

\subsection{DPM in Visual Domain}
Recent studies have shown that representations learned by DPM also capture high-level semantic information. Feature maps extracted in the later stages of the reverse diffusion process contain rich representations and are highly effective for segmentation tasks \cite{croitoru2023diffusion}. In the field of medical image segmentation, DPM has achieved new state-of-the-art (SOTA) results on several benchmark datasets \cite{kazerouni2023diffusion}.
DPM, based on probabilistic modeling, generates images progressively without the need for an adversarial discriminator, addressing some limitations of GAN and gaining popularity in various applications. Studies such as \cite{wolleb2022diffusion}, \cite{wyatt2022anoddpm}, and \cite{zhao2024dtan} have applied DPM to medical segmentation. However, the gradual denoising process can lead to the loss of fine details and challenges in maintaining global consistency. 

Hence, effectively utilizing useful information from the original image to guide the reverse diffusion process has become a critical challenge. \cite{amit2021segdiff} and \cite{chowdary2023diffusion} incorporate original image features to preserve details.
However, they do not fully utilize global features or address the complexities of feature fusion, particularly in challenging tasks like crack segmentation.
Meanwhile, \cite{wu2024medsegdiffv1} and \cite{wu2024medsegdiff} incorporate frequency domain guidance, which adds computational complexity and lacks generalization ability across different image types, especially heterogeneous data like fused grayscale and depth images.


\subsection{Grayscale and Depth Image Fusion in Defect Detection}
Fusing grayscale and depth images for defect detection presents several challenges, primarily due to the complementary yet distinct nature of the data. 
Grayscale images are sensitive to lighting and contrast variations, while depth images capture structural information but are often affected by significant background noise from pavements. 
The key challenge lies in effectively integrating these data sources, minimizing noise, and maximizing the complementary information each image provides.
Studies such as \cite{guan2021automated}, \cite{li2024cnn}, \cite{zhou2021crack} have fused grayscale and depth images, leveraging cross-domain feature correlations to achieve more comprehensive defect detection.
Despite these advancements, challenges remain in reducing noise and maintaining consistency across the fused data. Diffusion models offer promising potential in this area, as their progressive generation process enables better integration of multi-source information.

\begin{figure}[htb]
\centering{\includegraphics[width=0.5\textwidth]{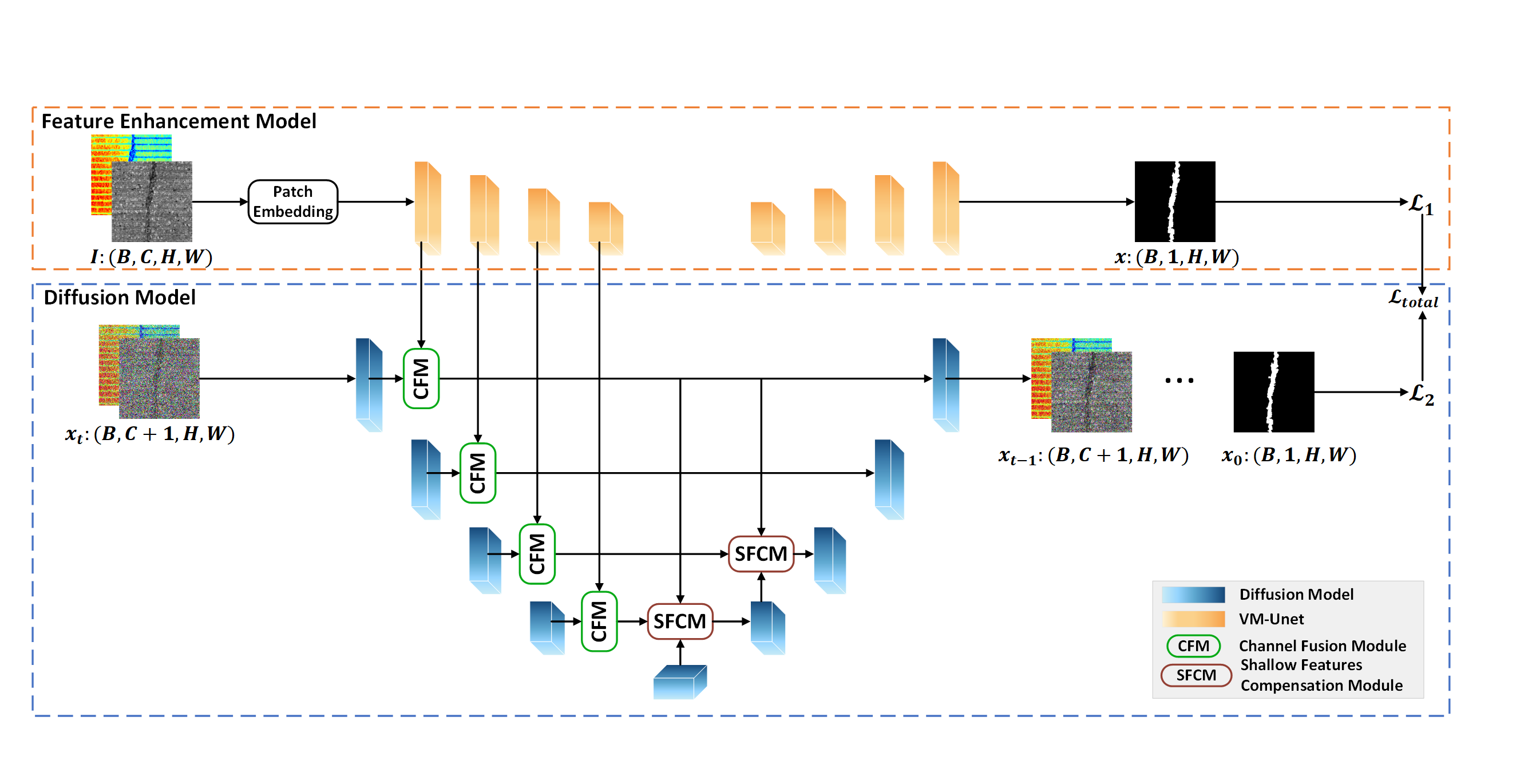}%
\label{fig_first_case}}
\caption{The overall architecture of CrackSegDiff. 
}
\label{fig_sim}
\end{figure}

\section{The Proposed Method}
As shown in Fig. 2, our framework builds upon SegDiff  (blue dashed region) based on DPM (\ref{subs:dpm}) and incorporates Feature Enhancement Model (FEM) to address the challenges in the crack segmentation field. 
The two modules are highly integrated by a Channel Fusion Module (CFM) in \ref{subs:cfm} and a Shallow Feature Compensation Module (SFCM) in \ref{subs:sfcm}.


The interaction between the FEM and the diffusion model is essential for ensuring the effectiveness of the incorporated global features. This crucial interaction is mediated by the loss function, which is designed to optimize both the integration of features and the diffusion process. As a result, this strategic control significantly enhances the accuracy of crack segmentation through improved feature augmentation.


\subsection{Diffusion Process of CrackSegDiff}
\label{subs:dpm}
DPM is a generative model parameterized by a Markov chain, consisting of a diffusion process of noises and a reverse diffusion process of reconstruction of original data.
The forward process defines the conditional distribution for each step of the reverse diffusion, where each step of the forward process guides the reverse diffusion. 
Specifically, it is used to define the conditional probability distribution of each step of the reverse diffusion, thereby guiding the model on how to progressively denoise and restore the original data. 

\textbf{Diffusion Process.} 
Given an initial crack data distribution $\mathrm{x}_{0}{\sim}q(x_{0})$, Gaussian noise \(\epsilon\sim N(0,1)\) can be continuously added to the crack segmentation mask $x_{t}$ at timestamp $t$ :
\begin{equation}
\label{equ:1}
x_t=\sqrt{\bar{\alpha_t}}x_{0}+\sqrt{1-\bar{\alpha_t}}\epsilon,~
{\bar{\alpha}}_{t}=\prod_{s=0}^{t}\alpha_{s}~,
\end{equation}
where \(\alpha_{t} = 1- \beta_t \) . 
Equ.\ref{equ:1} indicates that the standard deviation of $x_{t}$ is determined by a fixed value $\beta_{t}$. 
The state prediction is modelled as a Markov chain.
As $\text{t}$ increases, the final data distribution $x_T$ becomes an isotropic Gaussian distribution:
\begin{equation}
q(x_t|x_{t-1}):=N\big(x_t;\sqrt{1-\beta_t}x_{t-1},\beta_t\mathbf{I}\big),\beta_t\in(0,1)~,
\end{equation}
where $\mathbf{I}$ is the identity matrix.
The derivation of ${q}(x_t)$ at any moment can also be entirely based on $x_{0}$ and $\beta_{t}$ without iteration: 
 \begin{equation}q(x_t|x_0)=N\big(x_t;\sqrt{\bar{\alpha}_t}x_0,(1-\bar{\alpha}_t)\mathbf{I}\big)~.
 \end{equation}
 
\textbf{Reverse Diffusion Process.} 
This process involves recovering the original crack image from Gaussian noise. 
Assume the reverse diffusion process is also a Gaussian distribution, it is necessary to construct a parameter distribution function $p_{\theta}(x_{t-1}|x_{t})$ for estimation, as it is not feasible to fit the data distribution incrementally.
The reverse diffusion process is a Markov chain process: 
 \begin{equation}
 p_\theta(x_{t-1}|x_t)=N\left(x_{t-1};\mu_\theta(x_t,t),\sum\theta\left(x_t,t\right)\right)~.
 \end{equation}
 where $\theta$ represents the parameters of the reverse process. The key to the reverse diffusion process is designing an effective denoising network to predict the unknown $x_{0}$, with the known input $x_{t}$ and the time encoding $\text{t}$.

\subsection{CrackSegDiff Overview}

We model crack image segmentation as a discrete data generation task, aiming to estimate a segmentation map rather than noise. 
The input consists of grayscale and depth images, along with an additional noise channel, and the output is a refined segmentation map. 
According to the posterior distribution mean $\tilde{\mu}_{t}({x}_{t},{x}_{0})$ \cite{amit2021segdiff} in the forward diffusion process
\begin{equation}\begin{split}
\tilde{\mu}_{t}({x}_{t},{x}_{0})&=\frac{\frac{\sqrt{\alpha_{t}}}{\beta_{t}}{x}_{t}+\frac{\sqrt{\bar{\alpha}_{t}}}{1-\bar{\alpha}_{t}}{x}_{0}}{\frac{\alpha_{t}}{\beta_{t}}+\frac{1}{1-\bar{\alpha}_{t-1}}}\\&=\frac{\sqrt{\alpha_t}(1-\tilde{\alpha}_{t-1})}{1-\bar{\alpha}_t}{x}_t+\frac{\sqrt{\bar{\alpha}_{t-1}}\beta_t}{1-\bar{\alpha}_t}{x}_0\end{split} ~,
\end{equation}
since $x_{0}$ is unknown to the network, the proposed model predicts the segmentation map $x_{0}$ directly instead of the noise.


Multi-scale features allow the model to detect both small, fine cracks and larger structural patterns, ensuring detailed and robust segmentation across varying crack sizes and complex backgrounds. To better introduce crack image features, we extract multi-scale global features $F_g:[\mathbb{R}^{B\times i{C}\times\frac H{2^i}\times\frac W{2^i}}]_{i=1}^5$ (where ${i}$ is the scale and $B$ is the batch size) from the crack data through a FEM's encoder of the same size as the diffusion model encoder. 
Simultaneously, given the crack image data $I:\mathbb{R}^{B\times iC\times H\times W}$. 
The crack original image data and the noise are concatenated along the channel dimension as $x_{t}$, which is input to the encoder of the diffusion model to obtain the multi-scale local features $F_l:[\mathbb{R}^{B\times i\text{C}\times\frac W{2^i}\times\frac H{2^i}}]_{i=1}^5$.

Since $F_g$ and $F_l$ contain the same number and size of features, we merge features of the corresponding scales through the feature fusion module (CFM, details in \ref{subs:cfm}) to obtain fused features. 
Subsequently, the fused multi-scale features are supplemented with low-level features through the shallow feature compensation module (SFCM, details in \ref{subs:sfcm}) and input into the decoder of the diffusion model. 

We calculate $x_{t-1}$ by:
\begin{equation}\begin{split}
x_{t-1}=\alpha_t^{-\frac12}\Bigg(x_t-\frac{1-\alpha_t}{\sqrt{1-\bar{\alpha}_t}}\epsilon_\theta(concat(\boldsymbol{I},x_t),\boldsymbol{I},t)\Bigg)
\\+\vmathbb{1}_{[t>1]}\tilde{\beta}_t^{\frac12}z,z\sim N(0,\mathbf{I}),\tilde{\beta}_t=\frac{1-\bar{\alpha}_{t-1}}{1-\bar{\alpha}_t}\beta_t\end{split}\end{equation}
to obtain the predicted segmentation map $\mathrm{x}_{0}\in\mathbb{R}^{B\times C\times W\times H}$. When $t>1$, $\vmathbb{1}_{[t>1]}=1$.

Finally, the following gradient descent is used until the network converges: 
\begin{equation}\nabla_\theta||\epsilon-\epsilon_\theta(concat(\mathbf{I},x_t),\mathbf{I},t)||~.
\end{equation}

\subsection{Channel Fusion Module (CFM)}
\label{subs:cfm}
So far, while the diffusion model and FEM extract rich local and global features, respectively, the network struggles to fully utilize the spatial co-registration of grayscale and depth/range images. To address this, we introduce a channel fusion module to effectively balance and integrate these features, as shown in Fig. 3.
\begin{figure}[h]
\centering{\includegraphics[scale=0.33]{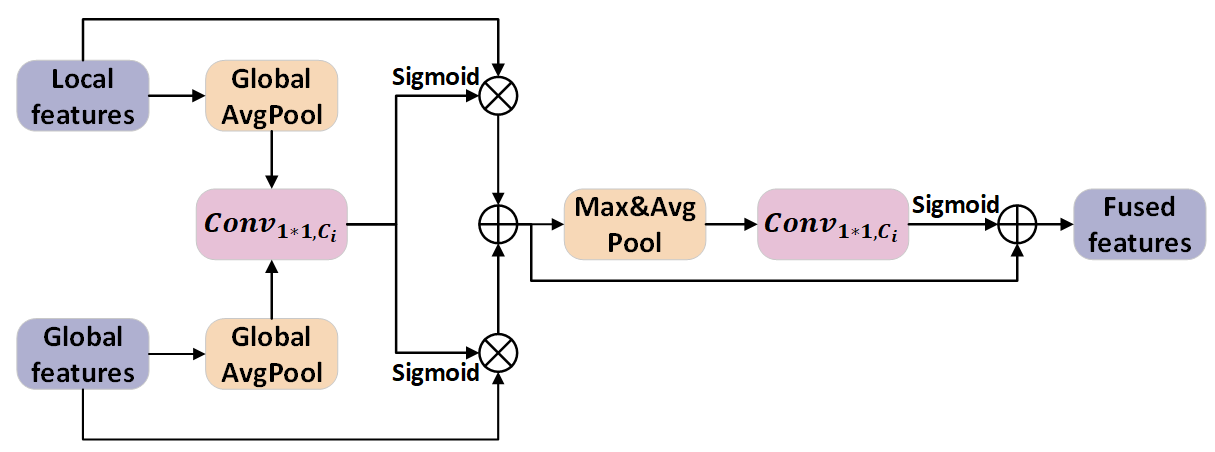}%
\label{fig_second_case}}
\caption{The overview of CFM.}
\label{fig_sim}
\end{figure}

As previous section, denote the local and global features extracted by the network as $F_l$ and $F_g$ , respectively. First, given that the input channel size of the i-th CFM module is $C_{i}$, these features are used for multi-scale feature extraction. The diffusion model extracts local features, while the FEM extracts global features, which are then aggregated through a global pooling layer. This allows the diffusion model to initialize with a rough but static global reference, helping to reduce diffusion variance.

Subsequently, a convolution with a kernel size of 1*1 is applied, producing two sets of weights through two different sigmoid functions. These weights are then multiplied element-wise with the local and global features, respectively, to highlight important features and suppress less important ones, thereby enhancing feature representation capability. The resulting features $F_m$ are then fused through the element-wise addition:
\begin{equation}\begin{split}F_m=F_g*Sigmoid(Conv_{1*1,C_i}(AvgPool(F_g))+\\F_l*Sigmoid(Conv_{1*1,C_i}(AvgPool(F_l))\end{split}~.
\end{equation}
Next, pooling and convolution operations aggregate effective features from the spatial dimension, maximizing the utilization of the spatial co-registration features of grayscale and depth images, yielding the final fused features $F_m{'}$:
\begin{equation}\begin{split}F_m{'}=F_m*Sigmoid(Conv_{1*1,C_i}(AvgPool(F_m)+\\MaxPool(F_m))\end{split}~.
\end{equation}

\begin{figure}[h]
\centering{\includegraphics[width=0.48\textwidth]{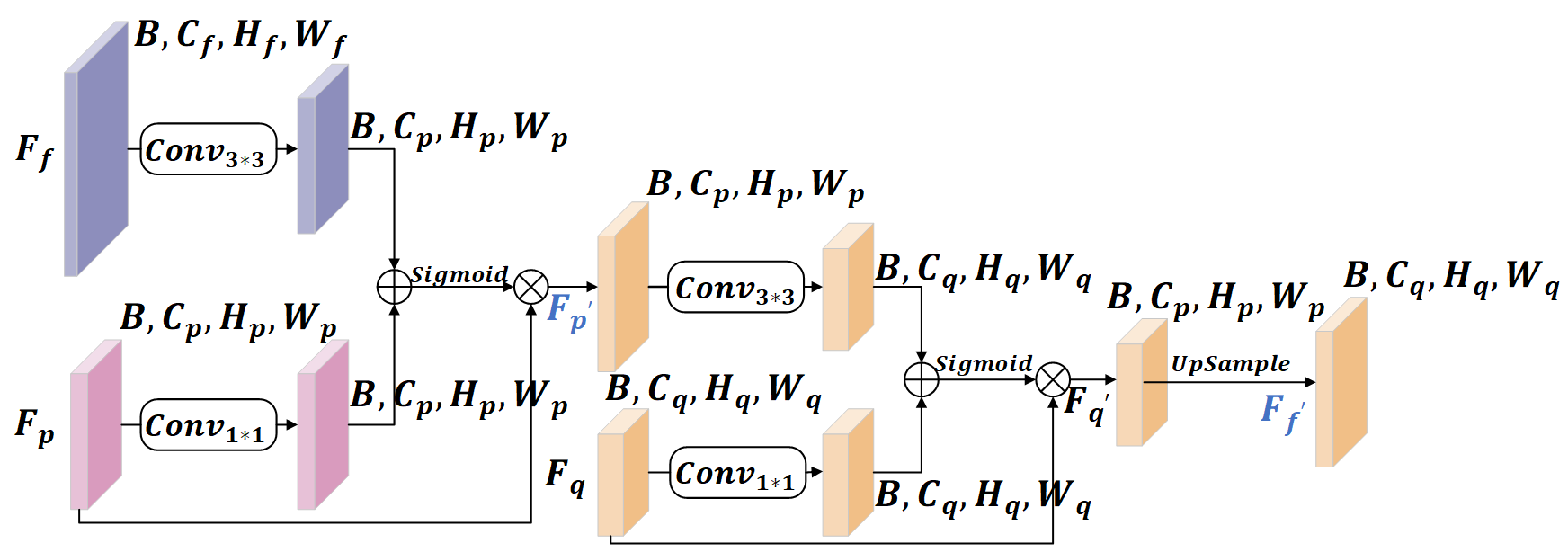}%
\label{fig_third_case}}
\caption{The overview of SFCM.}
\label{fig_sim}
\end{figure}
\subsection{Shallow Feature Compensation Module (SFCM)}
\label{subs:sfcm}
We observed that as network depth increases, some essential low-level features are lost. These features, derived directly from the original image data, are fundamental as they pertain to the physical properties of cracks, encapsulating crucial structural and textural information. The loss of these low-level features poses significant challenges in detecting shallow cracks. Given their high noise levels and minor representation in the overall feature set, incorporating these features into deeper network layers may seem beneficial. However, this integration can also introduce additional noise, complicating the prediction of shallow cracks.

To solve this issue, we propose a Shallow Feature Compensation Module to leverage the excellent noise resistance capabilities of diffusion model, as shown in Fig. 4.
\begin{equation}F_{p^{\prime}}=F_{p}*Sigmoid(Conv_{3*3}(F_{f})+Conv_{1*1}(F_{p}))~,
\end{equation}
\begin{equation}F_{f^{\prime}}=F_{q}*Sigmoid(Conv_{3*3}(F_{p^{\prime}})+Conv_{1*1}(F_{q}))~.
\end{equation}
SFCM integrates three distinct feature maps as inputs: $F_{f}$ from the initial layer following the fusion of the diffusion model and FEM, $F_{p}$ from the ($n-1$)-th skip connection, and $F_{q}$ from the $n$-th layer decoder. This integration is vital for accurately segmenting shallow cracks, which tend to lose detail in deeper layers. $F_{f}$ enhances both $F_{p}$ and $F_{q}$ by adding a substantial quantity of low-level features. This enrichment process culminates in the creation of the intermediate fused feature map $F_{p^{\prime}}$, and ultimately, the final fused feature map $F_{f^{\prime}}$.

SFCM enriches and supplements high-level features with multi-scale shallow features while retaining the original features, ensuring that detail-oriented information like shallow cracks does not weaken and disappear in deeper layers of the network. This effectively enhances low-level feature representation. Additionally, supported by the noise-resistant properties of the diffusion model, the module is less susceptible to extra noise in low-level features, thus improving the detection performance of shallow cracks.

\subsection{Loss function}
To ensure that SFM and SFCM inject effective features from FEM for the diffusion model, CrackSegDiff is trained using the total loss $\mathcal{L}_{total}$, which combines the classical diffusion noise prediction MSE loss $\mathcal{L}_{1}$ and the supervised feature supplementation network loss $\mathcal{L}_{2}$:
\begin{equation}
\mathcal{L}_1=\mathcal{L}_{mse}(\widehat{x_0},x_0)~,
\end{equation}
\begin{equation}\mathcal{L}_{2}=\mathcal{L}_{dice}(\widehat{x_{0}},x_{0})+\mathcal{L}_{bce}(\widehat{x_{0}},x_{0})~,
\end{equation}
\begin{equation}
\label{equ:14}
\mathcal{L}_{total}=\alpha\mathcal{L}_1+\beta\mathcal{L}_2~,
\end{equation}
where $\alpha$ and $\beta$ are set empirically to 1 and 10, respectively. 
These values balance the noise prediction loss and supervised feature extraction, ensuring effective feature learning while mitigating noise influence in crack segmentation tasks.
\begin{table*}[t]
\caption{Comparison of CrackSegDiff with state-of-the-art grayscale and depth fused segmentors on the FIND Dataset.}
\centering
\resizebox{\textwidth}{!}{%
\begin{tabular}{|l|lll|lll|lll|}
\hline
& \multicolumn{3}{l|}{Raw intensity}         & \multicolumn{3}{l|}{Raw range}         & \multicolumn{3}{l|}{Fused raw image}         \\ \hline& \multicolumn{1}{l|}{F1 score}         & \multicolumn{1}{c|}{IoU}              & BF score         & \multicolumn{1}{l|}{F1 score}         & \multicolumn{1}{c|}{IoU}              & BF score         & \multicolumn{1}{l|}{F1 score}         & \multicolumn{1}{c|}{IoU}              & BF score         \\ \hline
DenseCrack \cite{mei2020multi}&
\multicolumn{1}{l|}{\textit{68.2\%}} & \multicolumn{1}{l|}{56.5\%}          & \textit{-} & \multicolumn{1}{l|}{78.4\%}          & \multicolumn{1}{l|}{65.3\%}          & -          & \multicolumn{1}{l|}{81.5\%}          & \multicolumn{1}{l|}{69.7\%}          & -          \\ \hline
SegNet-FCN\cite{chen2020pavement}&
\multicolumn{1}{l|}{\textit{75.0\%}} & \multicolumn{1}{l|}{63.4\%}          & \textit{-} & \multicolumn{1}{l|}{81.1\%}          & \multicolumn{1}{l|}{68.6\%}          & -          & \multicolumn{1}{l|}{84.0\%}          & \multicolumn{1}{l|}{72.9\%}          & -          \\ \hline
CrackFusionNet\cite{zhou2021crack}                                                                  & \multicolumn{1}{l|}{\textit{77.8\%}} & \multicolumn{1}{l|}{66.5\%}          & \textit{-} & \multicolumn{1}{l|}{82.6\%}          & \multicolumn{1}{l|}{71.3\%}          & -          & \multicolumn{1}{l|}{86.8\%}          & \multicolumn{1}{l|}{77.3\%}          & -          \\ \hline
Unet-fcn \cite{zhang2021research}                                                                   & \multicolumn{1}{l|}{80.57\%}          & \multicolumn{1}{l|}{71.25\%}          & 84.44\%          & \multicolumn{1}{l|}{84.86\%}          & \multicolumn{1}{l|}{74.69\%}          & 87.44\%          & \multicolumn{1}{l|}{89.84\%}          & \multicolumn{1}{l|}{82.53\%}          & 91.56\%          \\ \hline
HRNet-OCR    \cite{wang2020deep}                                                                & \multicolumn{1}{l|}{78.55\%}          & \multicolumn{1}{l|}{67.73\%}          & 85.13\%          & \multicolumn{1}{l|}{84.89\%}          & \multicolumn{1}{l|}{74.18\%}          & 89.47\%          & \multicolumn{1}{l|}{85.07\%}          & \multicolumn{1}{l|}{75.55\%}          & 90.05\%          \\ \hline
Crackmer \cite{wang2024dual}                                                                    & \multicolumn{1}{l|}{76.54\%}          & \multicolumn{1}{l|}{64.92\%}          & 81.48\%          & \multicolumn{1}{l|}{81.78\%}          & \multicolumn{1}{l|}{69.72\%}          & 84.79\%          & \multicolumn{1}{l|}{87.32\%}          & \multicolumn{1}{l|}{78.25\%}          & 89.93\%          \\ \hline
CT-CrackSeg \cite{tao2023convolutional}                                                               & \multicolumn{1}{l|}{\textit{83.55\%}} & \multicolumn{1}{l|}{74.39\%}          & \textit{88.61\%} & \multicolumn{1}{l|}{88.51\%}          & \multicolumn{1}{l|}{80.17\%}          & 91.85\%          & \multicolumn{1}{l|}{92.75\%}          & \multicolumn{1}{l|}{87.06\%}          & 95.03\%          \\ \hline
MedSegDiff \cite{wu2024medsegdiffv1}                                                             & \multicolumn{1}{l|}{83.05\%}          & \multicolumn{1}{l|}{\textit{74.61\%}} & 88.21\%          & \multicolumn{1}{l|}{\textit{90.87\%}} & \multicolumn{1}{l|}{\textit{83.70\%}} & \textit{92.98\%} & \multicolumn{1}{l|}{\textit{95.03\%}} & \multicolumn{1}{l|}{\textit{90.77\%}} & \textit{96.50\%} \\ \hline
\textbf{\begin{tabular}[c]{@{}l@{}}CrackSegDiff (Ours)\end{tabular}} & \multicolumn{1}{l|}{\textbf{84.59\%}} & \multicolumn{1}{l|}{\textbf{77.31\%}} & \textbf{89.23\%} & \multicolumn{1}{l|}{\textbf{92.18\%}} & \multicolumn{1}{l|}{\textbf{86.11\%}} & \textbf{93.71\%} & \multicolumn{1}{l|}{\textbf{95.58\%}} & \multicolumn{1}{l|}{\textbf{91.90\%}} & \textbf{96.63\%} \\ \hline
\end{tabular}%
}
\end{table*}

\begin{figure*}[t] %
\begin{minipage}[h]{0.3\linewidth} %
\centering
\includegraphics[width=2in, height=1.9in]{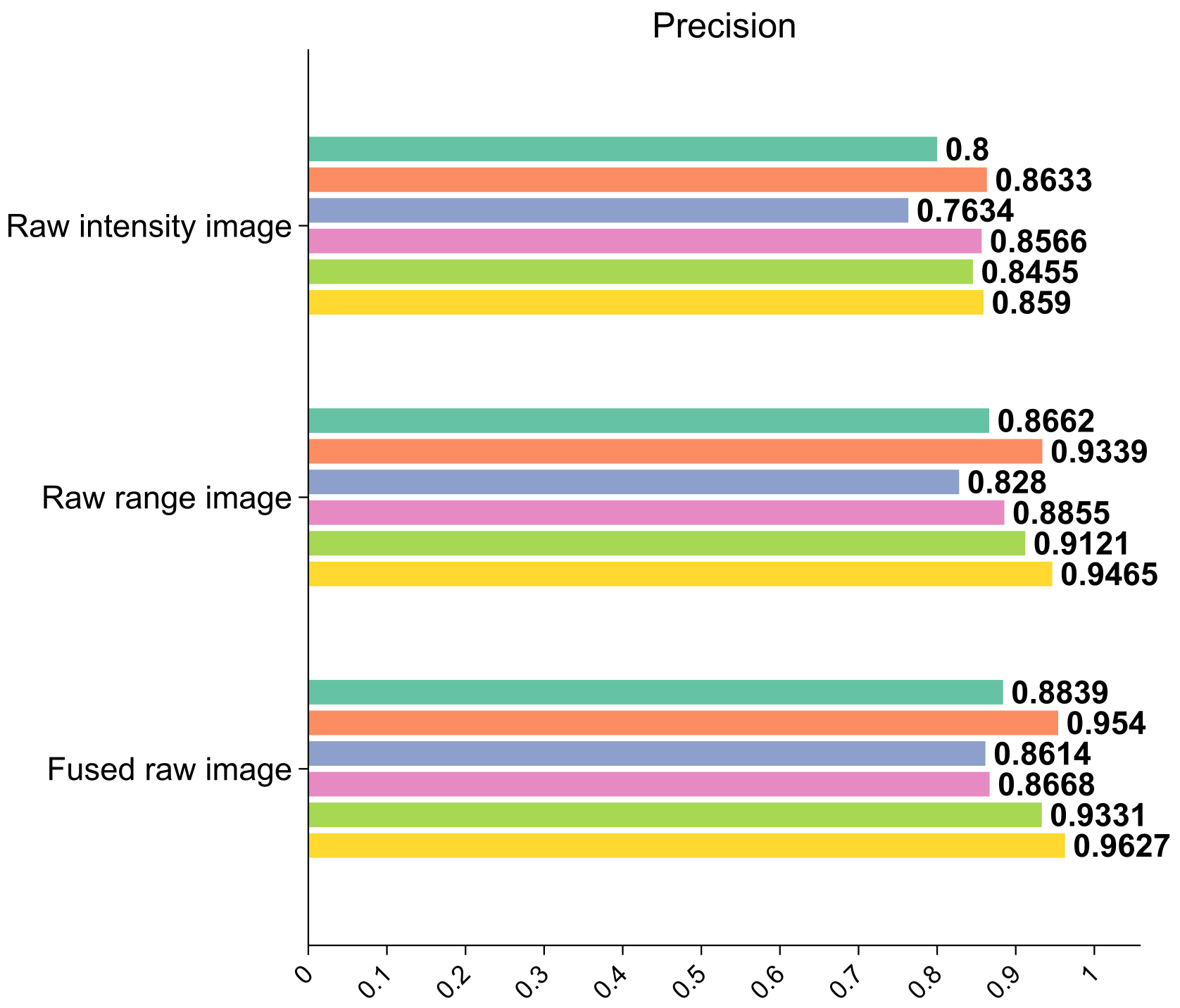} %
\end{minipage}%
\begin{minipage}[h]{0.3\linewidth}
\centering
\includegraphics[width=2in, height=1.9in]{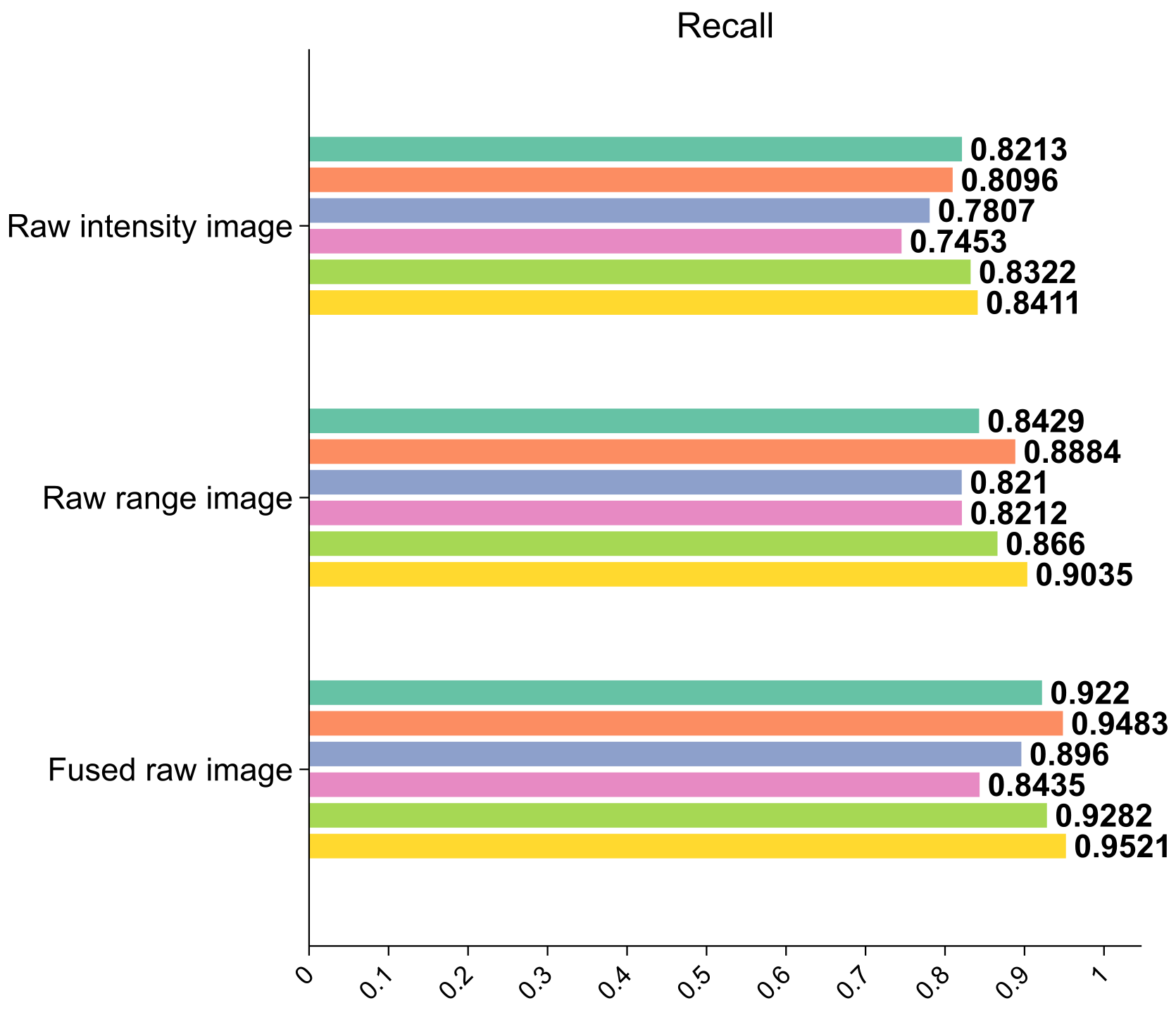}
\end{minipage}%
\begin{minipage}[h]{0.4\linewidth}
\centering
\includegraphics[width=2.5in, height=1.9in]{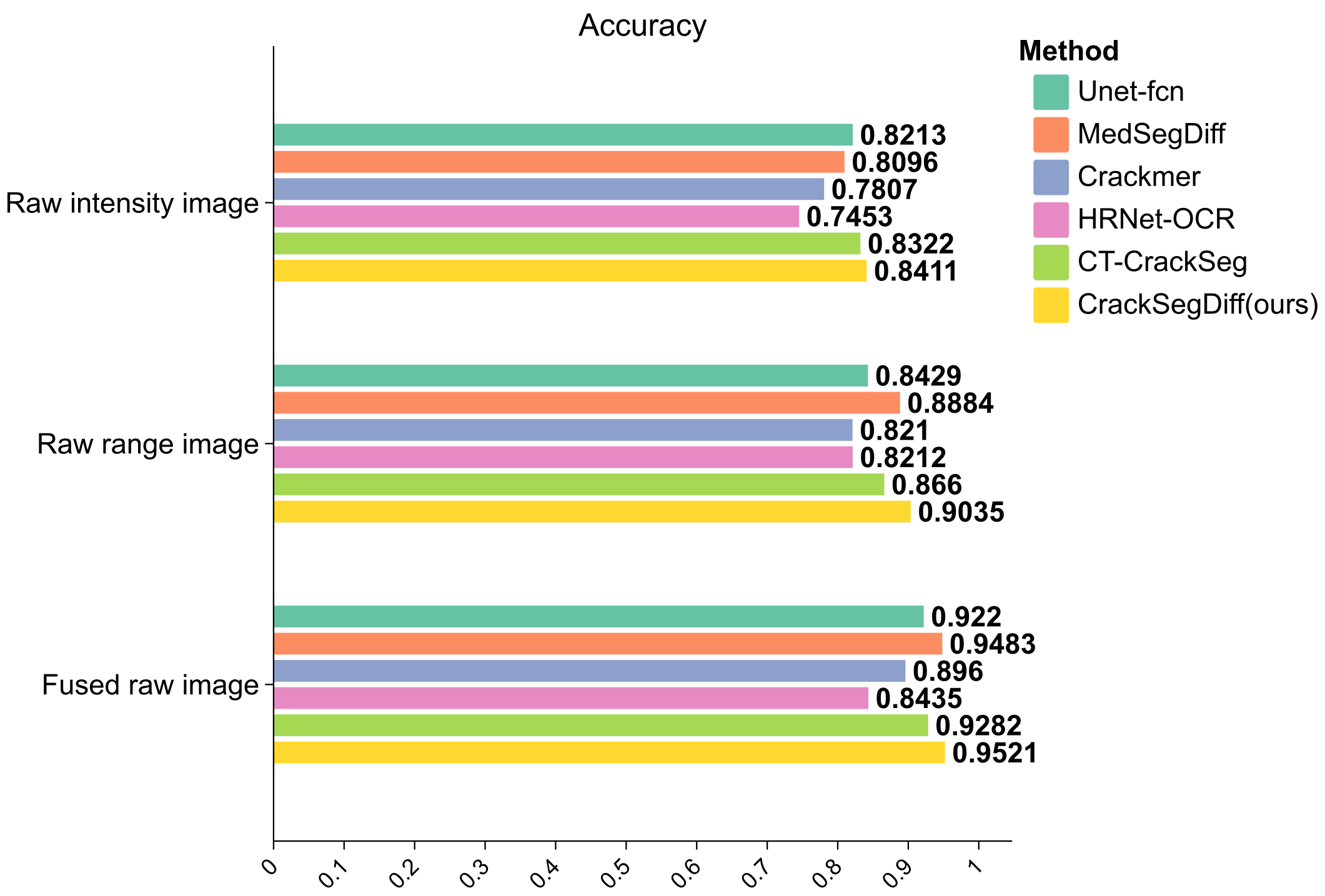}
\end{minipage}
\caption{Comparison charts of CrackSegDiff with state-of-the-art Segmentors on the FIND Dataset using precision, recall, and accuracy. 
}
\end{figure*}
\section{Experiments}
\subsection{Dataset}
We conduct experiments on the FIND dataset \cite{zhou2022fused}, which is \textit{currently the only publicly available dataset} to evaluate image fusion-based crack segmentors.
Moreover, since FIND captures data from multiple bridge decks and roads under real-world conditions, it is suitable to assess the models' robustness and generalization ability.
For one bridge deck or roadway region, FIND provides four data types, that is, grayscale images, range images, filtered range images, and fusion images, which are spatially registered and channel concatenated grayscale and range images. 
Each data type consists of 2,500 image patches with 256x256 pixels resolution and their corresponding pixel-level ground truth labels.

To verify the noise resistance capability of the proposed model, we did not apply any data pre-processing or augmentation to the original image data. 
Also, filtered range images in FIND are not used.
We randomly divided FIND into a training set and a testing set, containing 2,000 and 500 images, respectively. 

\subsection{Experimental Setup}
All experiments were conducted in PyTorch and trained and tested using a single NVIDIA A100 GPU. 
The initial learning rate of the network was set to $1\times10^{-4}$. The AdamW optimizer with a batch size of 8 was used to search for the optimal segmentation results over 200 epochs. 
For our model, we employed 1,000 diffusion steps, training in an end-to-end manner. 
For other methods, 500 images in the training set were used as the validation set. 
During training, predictions on the validation set were made every 2 epochs to avoid overfitting. 
In the final testing phase, all models were run once to obtain the final segmentation results. 

Segmentation performance was evaluated using the F1-Score \cite{fawcett2006introduction}, IoU \cite{mining2006introduction}, and BF-score \cite{csurka2013good} metrics.

\begin{table*}[t]
\caption{Ablation Study of modules in CrackSegDiff on the find dataset.}
\centering
\resizebox{\textwidth}{!}{%
\begin{tabular}{|l|lll|lll|lll|}
\hline
& \multicolumn{3}{l|}{Raw intensity}         & \multicolumn{3}{l|}{Raw range}         & \multicolumn{3}{l|}{Fused raw image}         \\ \hline         & \multicolumn{1}{l|}{F1 score}         & \multicolumn{1}{l|}{IoU}              & BF score         & \multicolumn{1}{l|}{F1 score}         & \multicolumn{1}{l|}{IoU}              & BF score         & \multicolumn{1}{l|}{F1 score}         & \multicolumn{1}{l|}{IoU}              & BF score         \\ \hline
SegDiff-\textit{baseline}             & \multicolumn{1}{l|}{81.95\%}          & \multicolumn{1}{l|}{71.69\%}          & 85.03\%          & \multicolumn{1}{l|}{87.46\%}          & \multicolumn{1}{l|}{78.89\%}          & 89.86\%          & \multicolumn{1}{l|}{91.94\%}          & \multicolumn{1}{l|}{84.72\%}          & 93.17\%          \\ \hline
+ SFCM                  & \multicolumn{1}{l|}{84.19\%}          & \multicolumn{1}{l|}{76.93\%}          & 88.77\%          & \multicolumn{1}{l|}{91.28\%}          & \multicolumn{1}{l|}{84.78\%}          & 93.26\%          & \multicolumn{1}{l|}{95.30\%}          & \multicolumn{1}{l|}{91.39\%}          & 96.54\%          \\ \hline
\textbf{+ SFM (Proposed)} & \multicolumn{1}{l|}{\textbf{84.59\%}} & \multicolumn{1}{l|}{\textbf{77.31\%}} & \textbf{89.23\%} & \multicolumn{1}{l|}{\textbf{92.18\%}} & \multicolumn{1}{l|}{\textbf{86.11\%}} & \textbf{93.71\%} & \multicolumn{1}{l|}{\textbf{95.58\%}} & \multicolumn{1}{l|}{\textbf{91.90\%}} & \textbf{96.63\%} \\ \hline
\end{tabular}%
}
\end{table*}

\subsection{Experimental Results}
We compare our method with the most advanced segmentation methods presented in the recent FIND dataset review \cite{zhou2023deep}, including Unet-fcn \cite{zhang2021research}, a prominent deep learning model for automatic pavement crack segmentation \cite{gong2023state}, and HRNet-OCR \cite{wang2020deep}, a well-regarded segmentation method in the field.
Additionally, we compared our model with other notable methods, including CNN and Transformer combined methods like Crackmer \cite{wang2024dual} and CT-CrackSeg \cite{tao2023convolutional}, and diffusion model-based MedSegDiff \cite{wu2024medsegdiffv1}.

\textbf{Quantitative evaluation.}
As shown in Table 1, CrackSegDiff ranks the first for all F1-Score, IoU, and BF-Score metrics across all three types of image data in the FIND dataset. 
Since the boundary distance thresholds of BF-score used in DenseCrack \cite{mei2020multi}, SegNet-FCN \cite{chen2020pavement}, and CrackFusionNet\cite{zhou2021crack} are unknown, for fair comparison, we do not compare their BF-scores.

Our method achieves more effective feature fusion than MedSegDiff. By incorporating global features and supplementing low-level details, it enhances crack structure and fine details, resulting in better performance.
As shown in Fig. 5, fused methods obtain better results than single source-based methods.
Our model outperforms other methods on all three metrics, that is, precision, recall, and accuracy. 

\textbf{Qualitative evaluation.}
As the upper part of Fig. 6 shows, Unet-fcn and HRNet-OCR suffer from severe noise interference, low-contrast or blurry regions, leading to incorrect predictions. 
Crackmer, MedSegDiff, and CT-CrackSeg fail to locate the correct crack positions. 
However, CrackSegDiff is not affected by noise interference and thus generates segmentation maps with precise details. 
Furthermore, HRNet-OCR and Crackmer perform poorly when using fused images, failing to effectively utilize the complementary sources.

As the lower sample in Fig. 6 shows, Unet-fcn and Crackmer are easily disturbed by shadows in grayscale images, while Unet-fcn, Crackmer, and MedSegDiff are affected by road grooves in range images. 
In contrast, our CrackSegDiff demonstrates exceptional noise resistance across all three types of image data.

\subsection{Ablation Study}
As shown in Table 2, the evaluations (F1-Score, IoU, and Bf-Score) are improved progressively as SFCM, and CFM modules are incorporated into the diffusion model. 
\begin{figure}[h]
\centering{\includegraphics[width=0.48\textwidth]{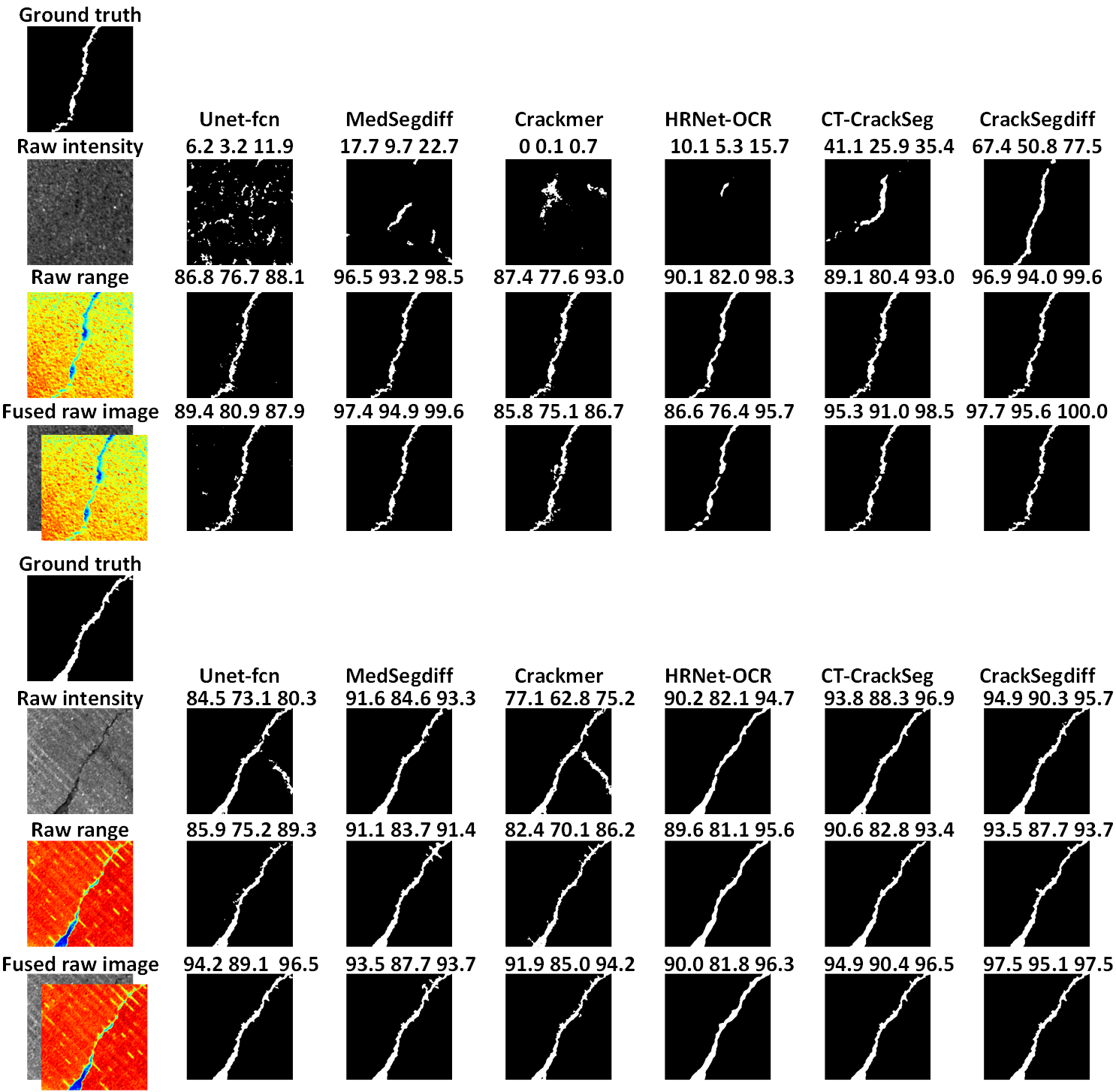}%
\label{fig_second_case}}
\caption{Qualitative comparison of CrackSegDiff with state-of-the-art segmentation methods. From left to right, the metrics used are F1-Score, IoU, and BF-Score.}
\label{fig_sim}
\end{figure}
This demonstrates that the SFCM and CFM can further enhance the crack segmentation accuracy of the diffusion model.

\textbf{Analysis of shallow cracks.}
The SFCM is specifically designed to address the challenge of recognizing shallow cracks, which is often difficult due to their limited context and subtle features. Although occurrences of such cracks are relatively infrequent in the test set, the impact of the SFCM module on overall performance metrics is not substantial, due to the small number of instances observed.

Despite the limited presence of shallow cracks in the test set, Figure 6 vividly shows the significant impact of the SFCM module. This improvement in a specific scenario underscores the module's effectiveness in enhancing shallow crack detection, suggesting that its benefits, while evident, may not be fully reflected in the overall aggregated performance metrics.

\textbf{Model illusion.}
Pavement cracks are often mistaken for plate joints, stains, shadows, and light reflections, which can lead to inaccuracies in model predictions. However, our model addresses this issue by employing fused images that combine spatially registered grayscale and depth data. This integration, alongside the model’s step-by-step denoising process, ensures alignment with the true data distribution, preserves realistic details, and minimizes the risk of generating artifacts or false hallucinations.

\section{Conclusion}
In this paper, we introduced a novel framework for crack image segmentation, CrackSegDiff, which leverages a DPM integrated with fused grayscale and range images.
To overcome the limitations of traditional DPM-based methods—specifically, their failure to capture global features and difficulties in fusing heterogeneous multi-source data—we developed two pivotal modules: CFM and SFCM. 
The CFM effectively integrates spatial registration features from both grayscale and range images, enhancing data coherence, while the SFCM specifically targets the improvement of shallow crack segmentation by restoring essential low-level features often lost in standard processing. 
Experimental results proves the enhanced capability of CrackSegDiff to adeptly manage complex crack segmentation tasks, demonstrating significant robustness against noise and background variations.
We hope this paper establishes a new benchmark in crack segmentation and also a new diagram for general DPM-based multi-modal image segmentation.

\bibliographystyle{IEEEtran}
\clearpage
\bibliography{ref}

\begin{thebibliography}{10}
\providecommand{\url}[1]{#1}
\csname url@samestyle\endcsname
\providecommand{\newblock}{\relax}
\providecommand{\bibinfo}[2]{#2}
\providecommand{\BIBentrySTDinterwordspacing}{\spaceskip=0pt\relax}
\providecommand{\BIBentryALTinterwordstretchfactor}{4}
\providecommand{\BIBentryALTinterwordspacing}{\spaceskip=\fontdimen2\font plus
\BIBentryALTinterwordstretchfactor\fontdimen3\font minus \fontdimen4\font\relax}
\providecommand{\BIBforeignlanguage}[2]{{%
\expandafter\ifx\csname l@#1\endcsname\relax
\typeout{** WARNING: IEEEtran.bst: No hyphenation pattern has been}%
\typeout{** loaded for the language `#1'. Using the pattern for}%
\typeout{** the default language instead.}%
\else
\language=\csname l@#1\endcsname
\fi
#2}}
\providecommand{\BIBdecl}{\relax}
\BIBdecl

\bibitem{amit2021segdiff}
T.~Amit, T.~Shaharbany, E.~Nachmani, and L.~Wolf, ``Segdiff: Image segmentation with diffusion probabilistic models,'' \emph{arXiv preprint arXiv:2112.00390}, 2021.

\bibitem{ruan2024vm}
J.~Ruan and S.~Xiang, ``Vm-unet: Vision mamba unet for medical image segmentation,'' \emph{arXiv preprint arXiv:2402.02491}, 2024.

\bibitem{mei2020multi}
Q.~Mei and M.~G{\"u}l, ``Multi-level feature fusion in densely connected deep-learning architecture and depth-first search for crack segmentation on images collected with smartphones,'' \emph{Structural Health Monitoring}, vol.~19, no.~6, pp. 1726--1744, 2020.

\bibitem{zhang2020crackgan}
K.~Zhang, Y.~Zhang, and H.-D. Cheng, ``Crackgan: Pavement crack detection using partially accurate ground truths based on generative adversarial learning,'' \emph{IEEE Transactions on Intelligent Transportation Systems (TITS)}, vol.~22, no.~2, pp. 1306--1319, 2020.

\bibitem{zhou2023hybrid}
Z.~Zhou, J.~Zhang, and C.~Gong, ``Hybrid semantic segmentation for tunnel lining cracks based on swin transformer and convolutional neural network,'' \emph{Computer-Aided Civil and Infrastructure Engineering}, vol.~38, no.~17, pp. 2491--2510, 2023.

\bibitem{vaswani2017attention}
A.~Vaswani, N.~Shazeer, N.~Parmar, J.~Uszkoreit, L.~Jones, A.~N. Gomez, {\L}.~Kaiser, and I.~Polosukhin, ``Attention is all you need,'' \emph{Advances in neural information processing systems}, vol.~30, 2017.

\bibitem{zhou2023deep}
S.~Zhou, C.~Canchila, and W.~Song, ``Deep learning-based crack segmentation for civil infrastructure: Data types, architectures, and benchmarked performance,'' \emph{Automation in Construction}, vol. 146, p. 104678, 2023.

\bibitem{wu2024medsegdiff}
J.~Wu, W.~Ji, H.~Fu, M.~Xu, Y.~Jin, and Y.~Xu, ``Medsegdiff-v2: Diffusion-based medical image segmentation with transformer,'' in \emph{AAAI Conference on Artificial Intelligence (AAAI)}, vol.~38, no.~6, 2024, pp. 6030--6038.

\bibitem{chowdary2023diffusion}
G.~J. Chowdary and Z.~Yin, ``Diffusion transformer u-net for medical image segmentation,'' in \emph{International conference on medical image computing and computer-assisted intervention (MICCAI)}, 2023, pp. 622--631.

\bibitem{zhu2024diffswintr}
J.~Zhu, H.~Zhu, Z.~Jia, and P.~Ma, ``Diffswintr: A diffusion model using 3d swin transformer for brain tumor segmentation,'' \emph{International Journal of Imaging Systems and Technology}, vol.~34, no.~3, p. e23080, 2024.

\bibitem{liu2024transdiff}
X.~Liu, Y.~Zhao, S.~Wang, and J.~Wei, ``Transdiff: medical image segmentation method based on swin transformer with diffusion probabilistic model,'' \emph{Applied Intelligence}, vol.~54, no.~8, pp. 6543--6557, 2024.

\bibitem{gu2023mamba}
A.~Gu and T.~Dao, ``Mamba: Linear-time sequence modeling with selective state spaces,'' \emph{arXiv preprint arXiv:2312.00752}, 2023.

\bibitem{di2023u}
A.~Di~Benedetto, M.~Fiani, and L.~M. Gujski, ``U-net-based cnn architecture for road crack segmentation,'' \emph{Infrastructures}, vol.~8, no.~5, p.~90, 2023.

\bibitem{liu2019deepcrack}
Y.~Liu, J.~Yao, X.~Lu, R.~Xie, and L.~Li, ``Deepcrack: A deep hierarchical feature learning architecture for crack segmentation,'' \emph{Neurocomputing}, vol. 338, pp. 139--153, 2019.

\bibitem{fei2019pixel}
Y.~Fei, K.~C. Wang, A.~Zhang, C.~Chen, J.~Q. Li, Y.~Liu, G.~Yang, and B.~Li, ``Pixel-level cracking detection on 3d asphalt pavement images through deep-learning-based cracknet-v,'' \emph{IEEE Transactions on Intelligent Transportation Systems (TITS)}, vol.~21, no.~1, pp. 273--284, 2019.

\bibitem{xiang2023crack}
C.~Xiang, J.~Guo, R.~Cao, and L.~Deng, ``A crack-segmentation algorithm fusing transformers and convolutional neural networks for complex detection scenarios,'' \emph{Automation in Construction}, vol. 152, p. 104894, 2023.

\bibitem{wang2024swincrack}
C.~Wang, H.~Liu, X.~An, Z.~Gong, and F.~Deng, ``Swincrack: Pavement crack detection using convolutional swin-transformer network,'' \emph{Digital Signal Processing}, vol. 145, p. 104297, 2024.

\bibitem{quan2023crackvit}
J.~Quan, B.~Ge, and M.~Wang, ``Crackvit: a unified cnn-transformer model for pixel-level crack extraction,'' \emph{Neural Computing and Applications}, vol.~35, no.~15, pp. 10\,957--10\,973, 2023.

\bibitem{gao2019generative}
Z.~Gao, B.~Peng, T.~Li, and C.~Gou, ``Generative adversarial networks for road crack image segmentation,'' in \emph{International Joint Conference on Neural Networks (IJCNN)}, 2019, pp. 1--8.

\bibitem{9770464}
A.~Sekar and V.~Perumal, ``Cfc-gan: Forecasting road surface crack using forecasted crack generative adversarial network,'' \emph{IEEE Transactions on Intelligent Transportation Systems (TITS)}, vol.~23, no.~11, pp. 21\,378--21\,391, 2022.

\bibitem{tian2021new}
L.~Tian, Z.~Wang, W.~Liu, Y.~Cheng, F.~E. Alsaadi, and X.~Liu, ``A new gan-based approach to data augmentation and image segmentation for crack detection in thermal imaging tests,'' \emph{Cognitive Computation}, vol.~13, pp. 1263--1273, 2021.

\bibitem{pan2023automatic}
Z.~Pan, S.~L. Lau, X.~Yang, N.~Guo, and X.~Wang, ``Automatic pavement crack segmentation using a generative adversarial network (gan)-based convolutional neural network,'' \emph{Results in Engineering}, vol.~19, p. 101267, 2023.

\bibitem{croitoru2023diffusion}
F.-A. Croitoru, V.~Hondru, R.~T. Ionescu, and M.~Shah, ``Diffusion models in vision: A survey,'' \emph{IEEE Transactions on Pattern Analysis and Machine Intelligence (TPAMI)}, vol.~45, no.~9, pp. 10\,850--10\,869, 2023.

\bibitem{kazerouni2023diffusion}
A.~Kazerouni, E.~K. Aghdam, M.~Heidari, R.~Azad, M.~Fayyaz, I.~Hacihaliloglu, and D.~Merhof, ``Diffusion models in medical imaging: A comprehensive survey,'' \emph{Medical Image Analysis (MIA)}, vol.~88, p. 102846, 2023.

\bibitem{wolleb2022diffusion}
J.~Wolleb, F.~Bieder, R.~Sandk{\"u}hler, and P.~C. Cattin, ``Diffusion models for medical anomaly detection,'' in \emph{International Conference on Medical image computing and computer-assisted intervention (MICCAI)}, 2022, pp. 35--45.

\bibitem{wyatt2022anoddpm}
J.~Wyatt, A.~Leach, S.~M. Schmon, and C.~G. Willcocks, ``Anoddpm: Anomaly detection with denoising diffusion probabilistic models using simplex noise,'' in \emph{IEEE/CVF Conference on Computer Vision and Pattern Recognition (CVPR)}, 2022, pp. 650--656.

\bibitem{zhao2024dtan}
Y.~Zhao, J.~Li, L.~Ren, and Z.~Chen, ``Dtan: Diffusion-based text attention network for medical image segmentation,'' \emph{Computers in Biology and Medicine}, vol. 168, p. 107728, 2024.

\bibitem{wu2024medsegdiffv1}
J.~Wu, R.~Fu, H.~Fang, Y.~Zhang, Y.~Yang, H.~Xiong, H.~Liu, and Y.~Xu, ``Medsegdiff: Medical image segmentation with diffusion probabilistic model,'' in \emph{Medical Imaging with Deep Learning (MIDL)}, 2024, pp. 1623--1639.

\bibitem{guan2021automated}
J.~Guan, X.~Yang, L.~Ding, X.~Cheng, V.~C. Lee, and C.~Jin, ``Automated pixel-level pavement distress detection based on stereo vision and deep learning,'' \emph{Automation in Construction}, vol. 129, p. 103788, 2021.

\bibitem{li2024cnn}
P.~Li, B.~Zhou, C.~Wang, G.~Hu, Y.~Yan, R.~Guo, and H.~Xia, ``Cnn-based pavement defects detection using grey and depth images,'' \emph{Automation in Construction}, vol. 158, p. 105192, 2024.

\bibitem{zhou2021crack}
S.~Zhou and W.~Song, ``Crack segmentation through deep convolutional neural networks and heterogeneous image fusion,'' \emph{Automation in Construction}, vol. 125, p. 103605, 2021.

\bibitem{chen2020pavement}
T.~Chen, Z.~Cai, X.~Zhao, C.~Chen, X.~Liang, T.~Zou, and P.~Wang, ``Pavement crack detection and recognition using the architecture of segnet,'' \emph{Journal of Industrial Information Integration}, vol.~18, p. 100144, 2020.

\bibitem{zhang2021research}
L.~Zhang, J.~Shen, and B.~Zhu, ``A research on an improved unet-based concrete crack detection algorithm,'' \emph{Structural Health Monitoring}, vol.~20, no.~4, pp. 1864--1879, 2021.

\bibitem{wang2020deep}
J.~Wang, K.~Sun, T.~Cheng, B.~Jiang, C.~Deng, Y.~Zhao, D.~Liu, Y.~Mu, M.~Tan, X.~Wang \emph{et~al.}, ``Deep high-resolution representation learning for visual recognition,'' \emph{IEEE transactions on pattern analysis and machine intelligence (TPAMI)}, vol.~43, no.~10, pp. 3349--3364, 2020.

\bibitem{wang2024dual}
J.~Wang, Z.~Zeng, P.~K. Sharma, O.~Alfarraj, A.~Tolba, J.~Zhang, and L.~Wang, ``Dual-path network combining cnn and transformer for pavement crack segmentation,'' \emph{Automation in Construction}, vol. 158, p. 105217, 2024.

\bibitem{tao2023convolutional}
H.~Tao, B.~Liu, J.~Cui, and H.~Zhang, ``A convolutional-transformer network for crack segmentation with boundary awareness,'' in \emph{IEEE International Conference on Image Processing (ICIP)}, 2023, pp. 86--90.

\bibitem{zhou2022fused}
W.~S. S.~Zhou, C.~Canchila, ``Fused image dataset for convolutional neural network-based crack detection (find),'' \url{https://doi.org/10.5281/zenodo.6383044}, 2022.

\bibitem{fawcett2006introduction}
T.~Fawcett, ``An introduction to roc analysis,'' \emph{Pattern recognition letters}, vol.~27, no.~8, pp. 861--874, 2006.

\bibitem{mining2006introduction}
W.~I.~D. Mining, \emph{Introduction to data mining}, 2006.

\bibitem{csurka2013good}
G.~Csurka, D.~Larlus, F.~Perronnin, and F.~Meylan, ``What is a good evaluation measure for semantic segmentation?'' in \emph{British Machine Vision Conference (BMVC)}, vol.~27, no. 2013, 2013, pp. 10--5244.

\bibitem{gong2023state}
H.~Gong, L.~Liu, H.~Liang, Y.~Zhou, and L.~Cong, ``A state-of-the-art survey of deep learning models for automated pavement crack segmentation,'' \emph{International Journal of Transportation Science and Technology}, 2023.

\end{thebibliography}
\end{document}